\documentclass[runningheads]{llncs}

\usepackage{eccv}

\usepackage{eccvabbrv}

\usepackage{graphicx}
\usepackage{booktabs}
\usepackage{threeparttable}
\usepackage{tabularx}
\usepackage[accsupp]{axessibility}  

\usepackage{hyperref}

\usepackage{orcidlink}
\usepackage{comment}
\usepackage{caption}
\usepackage{float}
\usepackage{placeins}

\begin{document}

\title{Limited but consistent gains in adversarial robustness by co-training object recognition models with human EEG} 

\titlerunning{Adversarial Robustness Gains via EEG Co-Training}

\author{Manshan Guo\inst{1,2,4}\orcidlink{0000-0002-5506-6854} \and
Bhavin Choksi \inst{1,2}\orcidlink{0000-0002-6475-4149} \and
Sari Sadiya\inst{1,2,3}\orcidlink{0009-0005-7482-3274} \and \\
Alessandro T. Gifford\inst{4}\orcidlink{0000-0002-8923-9477} \and
Martina G. Vilas\inst{1,2,5}\orcidlink{0000-0002-1097-8534} \and \\
Radoslaw M. Cichy\inst{4}\thanks{jointly directed work}\orcidlink{0000-0003-4190-6071} \and
Gemma Roig\inst{1,2,\star}\orcidlink{0000-0002-6439-8076}
}

\authorrunning{Guo et al.}

\institute{Department of Computer Science, Goethe University, Frankfurt am Main, Germany \\ 
\email{\{m.guo,choksi,Saba-Sadiya,roignoguera\}@em.uni-frankfurt.de} \and 
The Hessian Center for Artificial Intelligence (hessian.AI), Darmstadt, Germany \and 
Frankfurt Institute for Advanced Studies (FIAS), Frankfurt, Germany \and 
Department of Education and Psychology, Freie Universität Berlin, Berlin, Germany \\
\email{\{nikiguo93,rmcichy\}@zedat.fu-berlin.de,alessandro.gifford@gmail.com} \and
Ernst Strüngmann Institute for Neuroscience, Frankfurt am Main, Germany \\
\email{martina.vilas@esi-frankfurt.de} 
}

\maketitle

\begin{abstract}
  In contrast to human vision, artificial neural networks (ANNs) remain relatively susceptible to adversarial attacks. To address this vulnerability, efforts have been made to transfer inductive bias from human brains to ANNs, often by training the ANN representations to match their biological counterparts. Previous works relied on brain data acquired in rodents or primates using 
  invasive techniques, from specific regions of the brain, under non-natural conditions (anesthetized animals), and with stimulus datasets lacking diversity and naturalness. In this work, we explored whether aligning model representations to human EEG responses to a rich set of real-world images increases robustness to ANNs. Specifically, we trained ResNet50-backbone models on a dual task of classification and EEG prediction; and evaluated their EEG prediction accuracy and robustness to adversarial attacks. We observed significant correlation between the networks' EEG prediction accuracy, often highest around 100 ms post stimulus onset, and their gains in adversarial robustness. Although effect size was limited, effects were  consistent across different random initializations and robust for architectural variants. We further teased apart the data from individual EEG channels and observed strongest contribution from electrodes in the parieto-occipital regions. The demonstrated utility of human EEG for such tasks opens up avenues for future efforts that scale to larger datasets under diverse stimuli conditions with the promise of stronger effects. 
\end{abstract}

\section{Introduction}
\label{sec:intro}

Despite the remarkable performance of artificial neural networks (ANNs) in object recognition \cite{NIPS2012_alexnet,he2016resnet,simonyan2014vgg}, ANNs are sensitive to small so-called adversarial perturbations in the inputs \cite{szegedy2013intriguing}. Since the initial discovery of this vulnerability, the field has moved rapidly to devise various adversarial defenses against it\cite{madry2017towards,wang2023adversarial,cohen2019certified}.

In contrast, human perception is robust to adversarial attacks that are detrimental for ANNs\cite{elsayed2018adversarial,zhou2019humans}. This inspires the idea that adding more bio-inspired elements into ANNs might help alleviate their sensitivity to adversarial attacks and make them more robust. These biological inductive biases often are architecture-based, optimization-based or both \cite{konkle2023cognitive,choksi2021predify,huang2019brain,luo2015foveation,dapello2020simulating}. Studies have also directly constrained the ANN representations with their biological counterparts, often using neural data from rodents or non-human primates under non-ecological conditions as regularizers\cite{Sinz2019EngineeringAL,safarani2021towards,li2019learning,nassar20201,pirlot2022improving,federer2020improved,dapello2022aligning,lu2024realnet}. These attempts have led to promising, but modest gains in the robustness of the resulting ANNs \cite{federer2020improved,pirlot2022improving,safarani2021towards}. Yet, the cost and the practical challenges associated with acquiring such data also limits the diversity of stimuli used, often restricting the images from small datasets like CIFAR in grayscale, even for approaches relying on fMRI brain data\cite{rakhimberdina2022strengthening}.

Thus, in this work, we used a large-scale EEG dataset collected on diverse real-world images with the aim to improve ANNs robustness to adversarial attacks. Specifically, we experimented with the EEG dataset collected by \cite{gifford2022large} on participants viewing images from the THINGS dataset\cite{hebart2019things}. We report improvements against adversarial attacks which, though modest, were observed consistently across different random initializations of various architectural variants. These robustness gains were positively correlated with the ability of the models to predict EEG at early time points. Interestingly, as already observed in previous work, we also report similar robustness gains when using shuffled versions of the EEG data. While our results do not give state-of-the-art robustness, they provide important pointers guiding future research. We further analyze the EEG dataset across individual channels to investigate any channel-specific effects. We observe that mid-level channels (PO7, PO3, POz, PO4, PO8), though not as well-predicted by the ANNs as the early channels (Oz, O1, O2), are better (positively) correlated with the gains in robustness.
To further facilitate investigations into these methods, we publicly provide the code to the broader scientific community\footnote{Code available at: \url{https://github.com/cvai-roig-lab/eeg_cotraining_robustness}}.

\section{Related work}

Various efforts have aimed to imbue deep neural networks (ANNs) with human-like cognitive abilities by training them using brain data. Khosla \etal\cite{khosla2022high} showcased that ANNs optimized to predict fMRI activity in the fusiform face area (FFA) and extrastriate body area (EBA) could detect `faces' and `bodies', despite lacking direct exposure to such images during training. Fu \etal\cite{10254274} illustrated that aligning CNN representations with human fMRI data can improve model's performance in video emotion recognition tasks.

Among these, some studies have focused specifically on improving the adversarial robustness of object recognition models using brain data. Li \etal~\cite{li2019learning} recorded mice's neural response in primary visual cortex (V1) with photon-scans and then, along with classification, use it to penalize the representations of ResNet18\cite{kriegeskorte2008representational},  resulting in reduced vulnerability to white-box adversarial perturbations and gaussian noise.

Federer \etal\cite{federer2020improved} used publicly available recordings from V1 in monkeys using micro-electrode arrays\cite{coen2015flexible} and similarly penalized representations of VGG with RSA, observing a brain-like response to white-box adversarial noise and label corruption. In a similar fashion, Safarani \etal\cite{safarani2021towards} first jointly trained VGG19 for classification and predicting neural data collected in monkeys' V1, enhancing the models' robustness against 14 image distortions. Besides, they showed that their co-trained models were sensitive to salient regions of objects, reminiscent of V1's role in detecting object borders and bottom-up map.

Pirlot \etal\cite{pirlot2022improving} argued that the concurrent methods used for penalizing the representations could be limited, and instead proposed a Deep Canonical Correlation Analysis (DCCA)-based regularization, observing a reduction of vulnerability against adversarial noise. Dapello \etal\cite{dapello2022aligning} employed Centered Kernel Alignments (CKA) to penalize representations, leveraging monkeys' data in Inferior temporal cortex (IT) to improve robustness against white-box attacks and align with human behavior error patterns. Based on the findings from Stringer et al\cite{stringer2019high} that, regardless of the input, the eigenspectrum of covariance matrix of the neural code (in mice's visual cortex) followed a power law, Nassar \etal\cite{nassar20201} inquired whether a similar constraint on the ANN representations might help in enhancing their robustness. They found that, when implemented on smaller convolutional neural networks, their proposed spectral regularization improved robustness against $L_\infty$-FGSM and Projected Gradient Descent attacks\cite{madry2017towards}.

Unlike ours, most studies have restricted themselves to using brain data collected from non-human animals using costly invasive techniques. We note that a contemporary study took a similar rationale to ours and introduced a multi-layer alignment framework to align ANN representations with human EEG\cite{lu2024realnet}. Using the same EEG dataset, they trained an additional multi-layer module to predict image categories and human EEG from CORnet-S features (pretrained on ImageNet). While they tested their networks on FGSM attacks, which are known to be quite weak and prone to gradient-masking\cite{tramer2020adaptive}, their main focus was to demonstrate the utility of the learned representations for neuroscience---to better explain fMRI and behavioral data. 

\begin{figure}[!tb]
  \centering
  \includegraphics[width=\linewidth]{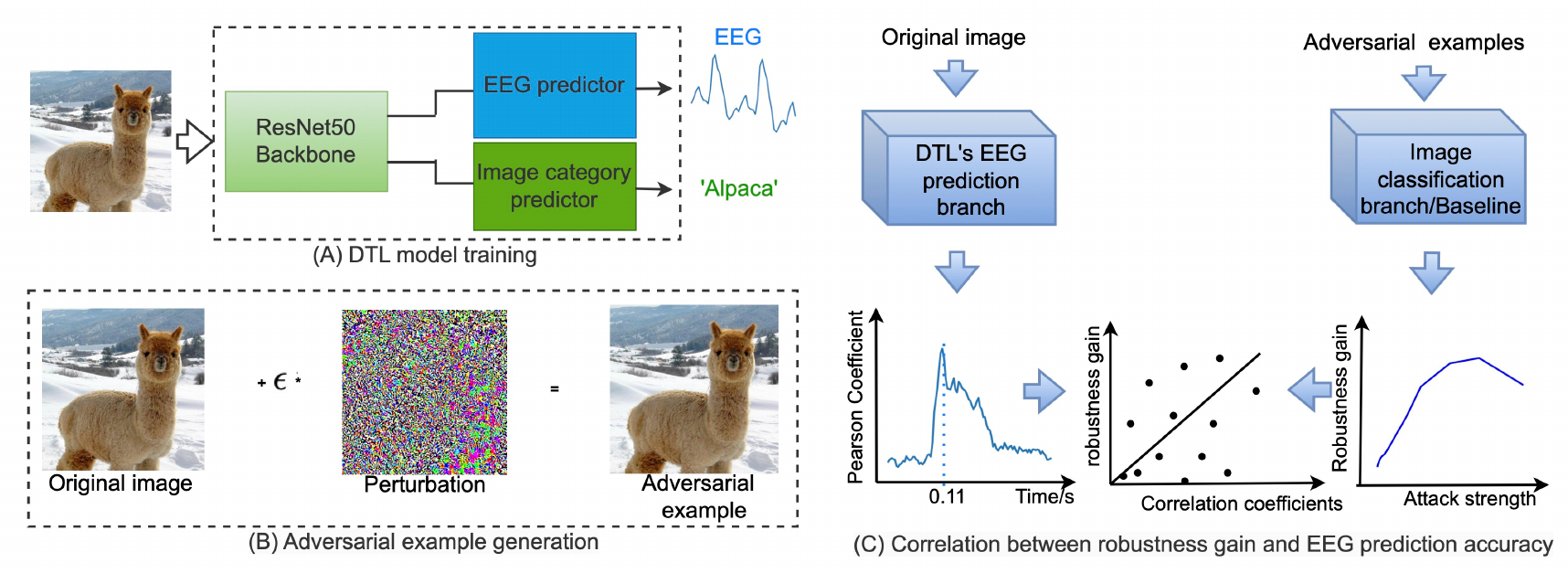}
  \caption{\textbf{Paradigm for improving adversarial robustness via co-training with human EEG:} We first trained dual-task learning (DTL) models with original and shuffled EEG data and then evaluated their robustness against various adversarial attacks. We trained four clusters of ResNet50 backbone models, each incorporating a different independent EEG predictor: Dense Layers (CNN), Recurrent Neural Networks (RNN), Transformer, and Attention layers. Finally, we measured the relationship between adversarial robustness gain and EEG prediction accuracy.}
  \label{fig:Paradigm}
\end{figure}

\section{Methods}

\paragraph{\textbf{Dataset}} We used a publicly available dataset containing images and corresponding EEG recordings from 10 subjects viewing images from the THINGS database \cite{gifford2022large, hebart2019things}. The training set included 16,540 natural images across 1,654 object categories, with each category containing 10 images presented in 4 separate runs. We split the training set into 9:1 split for training and validation, allocating 9 images per category for training (14,886 images total) and 1 image per category for validation (1,654 images total). The raw EEG signals were epoched from 200ms before to 800ms after stimulus onset and down-sampled to 100Hz. Seventeen channels were selected, including `Pz', `P3', `P7', `O1', `Oz', `O2', `P4', `P8', `P1', `P5', `PO7', `PO3', `POz', `PO4', `PO8', `P6', and `P2', which record signals from the occipital and parietal cortex where the visual signals are the strongest. Further details on EEG-image pair preprocessing are in \Cref{sec_app:distort}.

\paragraph{\textbf{Architectures and training}}

The DTL networks consisted of an image classification branch (a ResNet50 backbone) and an EEG prediction branch. Along with certain layers that were shared from the classification branch, the EEG prediction branch comprised of an independent component of either dense linear layers (CNN), recurrent layers (RNN), Transformer, or attention layers that were appended to the ResNet backbone. Overall 24 models (see \Cref{sec_app:arch&training} for additional details) were trained for two objectives---classification and EEG prediction. For the classification branch, 1654 object-categories were used. For EEG prediction, MSE loss was applied to predict the EEG data consisting of 100 timepoints (output of size  $1 \;image \times 17 \;channels \times 100 \;tps$). To balance the losses between these tasks, we applied the following total loss function from \cite{kendall2018multi}~:  

\begin{equation}
L(\textbf{W}, \delta_1, \delta_2) = \frac{1}{2 \delta_1^2} L_1(\textbf{W}) + \frac{1}{2 \delta_2^2} L_2(\textbf{W}) + log \delta_1 +log \delta_2.
\end{equation}

Here, $L_1$ and $L_2$ represent the EEG prediction and image classification losses, respectively, with $\frac{1}{2 \delta_1^2}$ and $\frac{1}{2 \delta_1^2}$ as loss coefficients. These parameters, along with the model weights \textbf{W} were updated using the Adam optimizer with a learning rate of 5e-6 and a weight decay of 0.0. The image classification branch was pre-trained on ImageNet, and the EEG prediction branch was initialized with three different training seeds (0, 17, and 337). Each model was trained for 200 epochs with a batch size of 64. As control experiments, we also trained the networks on three simulated EEG datasets obtained by shuffling the original, and randomly drawing from a geometrical or normal distribution. The models trained with such data are labeled as DTL-shuffled,  DTL-random, and DTL-random-normal and were contrasted with those trained on original EEG data (DTL-real). Additional details regarding all the architectures and model training can be found in~\Cref{sec_app:arch&training}.

\paragraph{\textbf{EEG prediction evaluation}}

Following \cite{gifford2022large}, we evaluated the EEG prediction results by measuring Pearson correlation between the predicted and the actual EEG. A PCC matrix (of shape $17 \times 100$) was constructed where each element represented the linear correlation between the predicted and actual EEG. We then averaged across the channel dimension to get a global value. Results were averaged across the 10 subjects and models initialized with three random seeds. Further details are available in \Cref{sec_app:pgd&eeg}.

\paragraph{\textbf{Adversarial robustness evaluation} and \textbf{Robustness gain}} Adversarial perturbations are image transformations capable of fooling ANNs while remaining imperceptible for humans. To assess the adversarial robustness of our models, we employed Foolbox \cite{rauber2017foolbox} to create adversarial versions of the 1654 original validation images under different attack strengths $\epsilon$. In particular, 1654 adversarial examples with each $\epsilon$ value were fed into each DTL model to obtain the the top-1 classification accuracy, denoted as ${acc_{DTL}(\epsilon)}$. Similarly, we obtained ${acc_{baseline}(\epsilon)}$ for the baseline model--the ResNet50 model trained for classification. The adversarial robustness gain was defined as $Gain_{DTL}(\epsilon) = acc_{DTL}(\epsilon)-$ $acc_{baseline}(\epsilon)$. We applied $L_2$-
and $L_\infty$-norm bounded untargeted projected gradient descent (PGD)~\cite{madry2017towards}, and $L_2$ Carlini \& Wagner (C\&W) attack~\cite{carlini2017evaluating}, as described in \Cref{sec_app:pgd&eeg}.

\paragraph{\textbf{Correlation between adversarial robustness gain and EEG prediction}}
We co-trained 24 architectures on EEG from 10 subjects using 3 training seeds, resulting in a total of  $24 \times 10 \times 3 = 720$ samples for correlation analysis. The mean adversarial robustness gain ${Avg\_Gain_{DTL}}$ was computed by averaging $Gain_{DTL}^n(\epsilon)$ across subjects and training seeds ($0 \leq n \leq 720$) as well as attack strength $\epsilon$. To identify significant time points (tps) contributing to robustness gain, we averaged ${PCC (ci,tps)}$ across all channels (ci, channel index) and time points (tps) within an optimal sliding window size, resulting in ${Avg\_PCC\_tps}$, which denotes mean prediction accuracy of all channels within the window. The sliding window size was optimized to encompass as many significant time points as possible, further details of which are provided in \Cref{sec_app:mean_gain_pcc}.

After identifying these significant time points, we investigated key electrodes by similarly measuring correlation between robustness gain and mean prediction accuracy per channel across critical time points.

\begin{figure}[!t]
  \centering
  \includegraphics[width=1.1\textwidth]{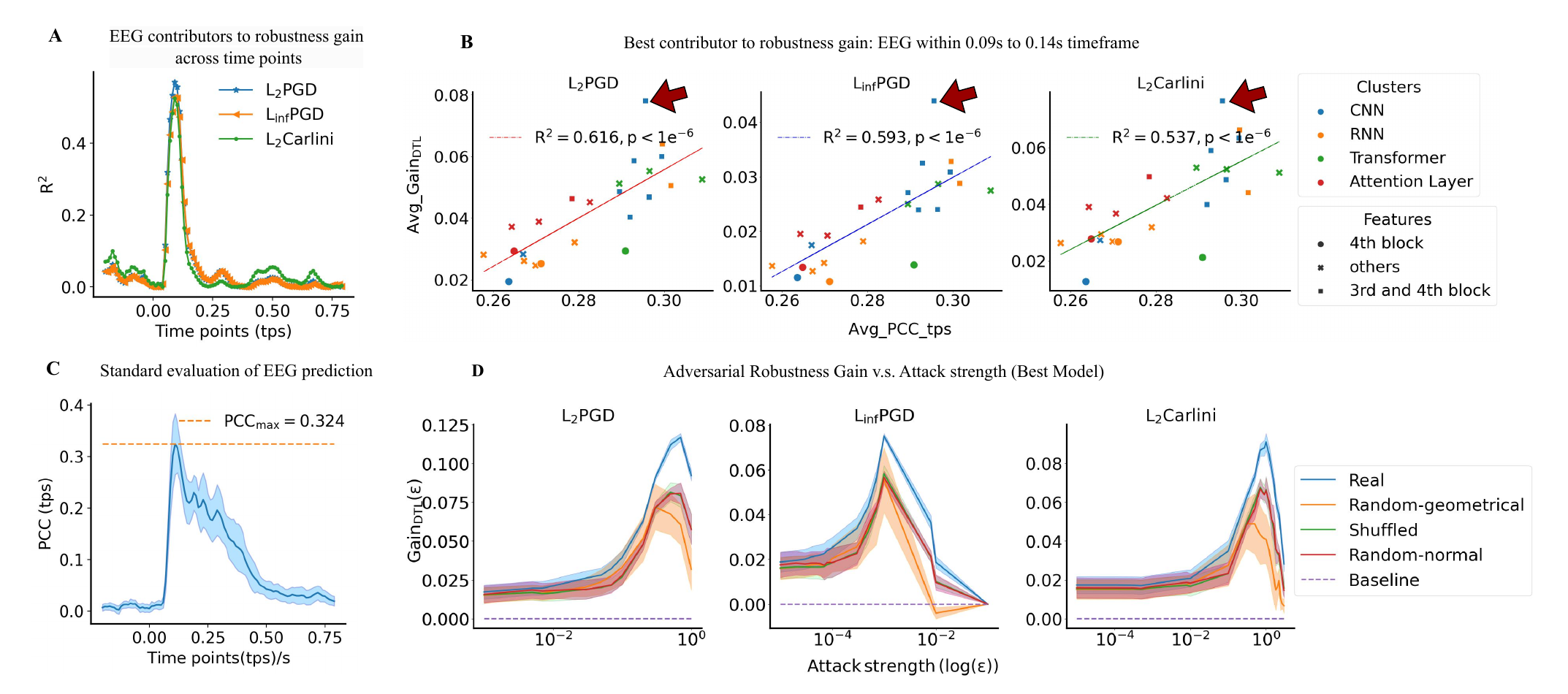}
  \caption{\textbf{Adversarial robustness gain was correlated with EEG prediction} (A) correlation value between mean adversarial robustness gain ($Avg\_Gain_{DTL}$) and mean EEG prediction accuracy ($Avg\_PCC\_tps$) peaked at 0.09s (B) Adversarial robustness gains across all the three attacks were significantly correlated with prediction accuracy of EEG from 0.09s to 0.14s. The correlation values $R^2$ correspond to the peak $R^2$ in (A). Colors denote architecture type and markers denote if the features used for EEG prediction were from only the 4th block, both the 3rd and 4th block (via concatenating/averaging of features from 3rd and 4th block, all 4 blocks, the last 3 blocks...), or other blocks excluding the 4th. Blue squares represent integrating 3rd and 4th block features for EEG prediction, which generally achieved higher $Avg\_Gain_{DTL}$ and $Avg\_PCC\_tps$. (C) EEG prediction of our most robust model (shown red arrow in B) using Pearson correlation coefficients. The correlation around 100ms is significant (p \textless 0.05, Bonferroni corrected). (D) Adversarial robustness gain ($Gain_{DTL}(\epsilon)$) of the model denoted in (B) along with the controls co-trained on shuffled and randomly generated EEG. Shaded regions represent the standard error over training seeds and subjects. ).}
  \label{fig:corr_slide}
\end{figure}

\section{Results}
\label{Results}

\paragraph{\textbf{Adversarial robustness gains were positively correlated with the models' EEG prediction}}

We first investigated if there is a relationship between the model's EEG prediction ability and its gains in adversarial robustness. For this, we measured the correlation between the gains in adversarial robustness ($Gain_{DTL}(\epsilon)$) of the 720 models considered here and their EEG prediction (AVG\_PCC\_tps); see Figure \ref{fig:corr_slide}A. We observed significant positive correlation values (between 0.53-0.61, p-values < 1e-6) implying that the robustness of the models scaled with the ability of the networks to better capture the statistics from the neural data. We note that similar results were obtained by \cite{dapello2020simulating} albeit for neural data from individual neurons in the macaque brain. Concerning the role of architecture, we observed that evaluations that combined the features from both the 3rd and 4th block obtained good results for both adversarial robustness and EEG prediction.

We further illustrate the robustness gains of our most robust model (in Figure~\ref{fig:corr_slide}B and D). This network (shown with a red arrow in Figure~\ref{fig:corr_slide}B) concatenated the features from both the 3rd and 4th block. We observed highest correlations of ~0.32 around 100 ms post-stimulus onset (Figure~\ref{fig:corr_slide}C), that is at the time of highest discriminability in the EEG data\cite{gifford2022large}. In Figure~\ref{fig:corr_slide}D, we show the robustness of this model against all the attacks used in our analysis. We also include the robustness gains obtained after training the model with the control (shuffled and random) versions of EEG . While these also showed some gains in robustness (as also reported in previous works), the model trained with the real EEG showed the highest robustness gains. While the gains are modest---a clear limitation of our results and similar to those reported in earlier studies---they are nonetheless surprising given the high strengths of the attacks (epsilon budget of 1. in L2 norm). Moreover, they demonstrate the potential for the utility of human EEG for rendering robustness. 

\begin{figure}[!t]
    \centering
    \includegraphics[width=\textwidth]{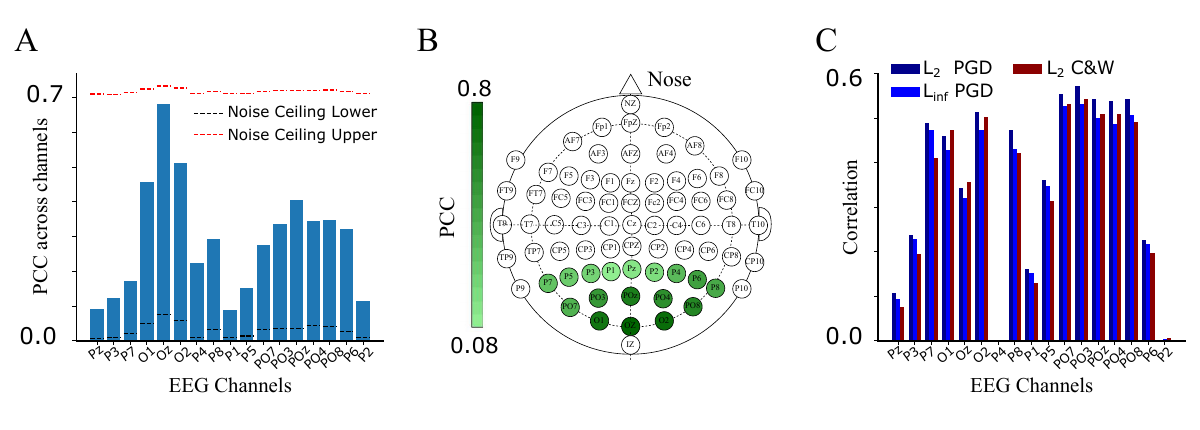}
    \caption{\textbf{Contribution of individual EEG electrodes in network robustness} (A) The PCC values across channels, calculated as the correlation between the predicted and the real EEG from 0.09s to 0.14s. The lower and upper noise ceilings are calculated as per \cite{gifford2022large} and are denoted in black and red lines respectively. (B) A 64-channel EEG top down brain view with the 17 electrodes covering occipital and parietal cortex colored with the PCC values obtained (from (A)) for each electrode. (C) Correlation values between the robustness gains (for each attack) and PCC values for each EEG channel }
    \label{fig:figure3}
\end{figure}

\paragraph{\textbf{Electrodes from mid-level EEG channels contribute most strongly to robustness}}

To find out which EEG channels mostly contributed to robustness, we first determined the EEG channels that were best predicted after our dual-task training. We measured the correlation between the original and predicted EEG, and observed that the models best predicted the data from early occipital channels (Oz, O1, O2) with the group of parieto-occipital channels (PO7, PO3, PO4, PO8, POz) being second best (\Cref{fig:figure3}A). These EEG channels overlay the visual cortex (see \Cref{fig:figure3}B), consistent with the origin of the observed signals\cite{gifford2022large}.

But do these channels actually contribute to the robustness of the networks? To ascertain that, we measured the correlations between the robustness gains for each attack and the PCC values for each individual channel (\Cref{fig:figure3}C). While the channels in the early visual areas seemed to be best predicted, those in the parieto-occipital region (from electrodes PO7, PO3, PO4, PO8, POz) showed the highest correlation values, indicating that it were the statistics from these channels that particularly aided in enhancing the robustness of the networks. This suggests that brain signal from the later visual processing stages in the human brain contribute more to the robustness than earlier processing stages.

\section{Discussion and Conclusion}
\label{sec:diss}

In this study, we explored the effectiveness of human EEG to render robustness to ANNs. Specifically, we co-trained the ANNs to predict human EEG signals in addition to image classification, and tested their robustness to adversarial perturbations. We observed consistent robustness gains across different variants of the networks. Our investigations revealed a positive correlation between a model's robustness and its ability to predict the EEG. We further teased apart the contribution from individual EEG channels, and observed that though the channels overlaying the early visual cortex were best predicted (with Oz even reaching the upper estimates of the noise ceilings), the ones in the parieto-occipital region correlated better with the gains in adversarial robustness.

Our work validates the use of human EEG data for enhancing the robustness of ANNs which, compared to intracranial recordings that were often used previously, is cheaper and easier to collect. Given that there is an ongoing trend in NeuroAI \cite{allen2022massive,lahner2023bold,cichy2019algonauts,CNeuroMod} to collect massive datasets, our methods could not only benefit from the new datasets, but also inform future data collection process. Future works could investigate if larger EEG datasets, collected with different stimulus conditions, say from the auditory domain, can similarly help in improving the robustness of artificial neural networks.

As reported in previous works, the robustness gain was consistent, yet modest. This could be because of the exact methods that we used to regularize our networks, as also suggested by \cite{pirlot2022improving}, or could be the inherent limitations of this approach in itself. Like earlier works, we found that the control (shuffled and random) versions of the EEG, and thus the mere statistics of the signal, also helped in rendering robustness to the ANNs. Indeed, this consistent observation, now observed across intra- and extracranial neural activity, deserves future scientific inquiry in its own right. Exactly what (statistical) elements of the neural activity are the networks utilizing to improve their robustness? Can we use these over conventional initialization methods to improve robustness of ANNs? These questions raise the need and guide future research efforts.

\section*{Acknowledgements}
This project was funded by the German Research Foundation (DFG) - DFG Research Unit FOR 5368 (GR) awarded to Gemma Roig, Deutsche Forschungsgemeinschaft (DFG; CI241/1-1, CI241/3-1, and CI241/7-1) awarded to Radoslaw Cichy, and a European Research Council (ERC) starting grant (ERC-2018-STG 803370) awarded to Radoslaw Cichy. We are grateful for access to the computing facilities of the Center for Scientific Computing at Goethe University and Freie universität Berlin. M. Guo is supported by a PhD stipend from the China Scholarship Council (CSC).

%
%
\bibliographystyle{splncs04}
\bibliography{main}

\newpage
\appendix
\section{Appendix}
\label{sec_app:methods}

\subsection{EEG-Images pairs}
\label{sec_app:distort}

\paragraph{EEG-Image pre-processing}
The raw EEG signals were first epoched into trials ranging from 200ms before the stimulus onset (denoted as -0.2s) to 800ms after the stimulus onset (denoted as 0.8s) and later down-sampled to 100Hz. 17 channels overlying occipital and parietal cortex where the visual signals are strongest were finally selected. Consequently, our EEG data matrix for model training is of shape ($14,886 \;images \times 4 \;trials \times 17 \;EEG \;channels \times 100 \;EEG \;tps (time \,points)$) and our EEG data matrix for model validation is of shape ($1654 \;images \times 4 \;trails \times 17 \;EEG \;channels \times 100 \;EEG \;tps$). As EEG signals are noisy, we averaged EEG data across the trial dimension and normalized it across the temporal dimension with Z-score. All the image stimuli were normalized and resized to $3 \times 224 \times 224 \;pixels$. 


\subsection{Dual task learning (DTL)}
\label{sec_app:arch&training}

\begin{figure}[!htb]
      \centering 
      \begin{subfigure}[t]{0.9\textwidth}
        \includegraphics[width=10.5cm,height=4.5cm]{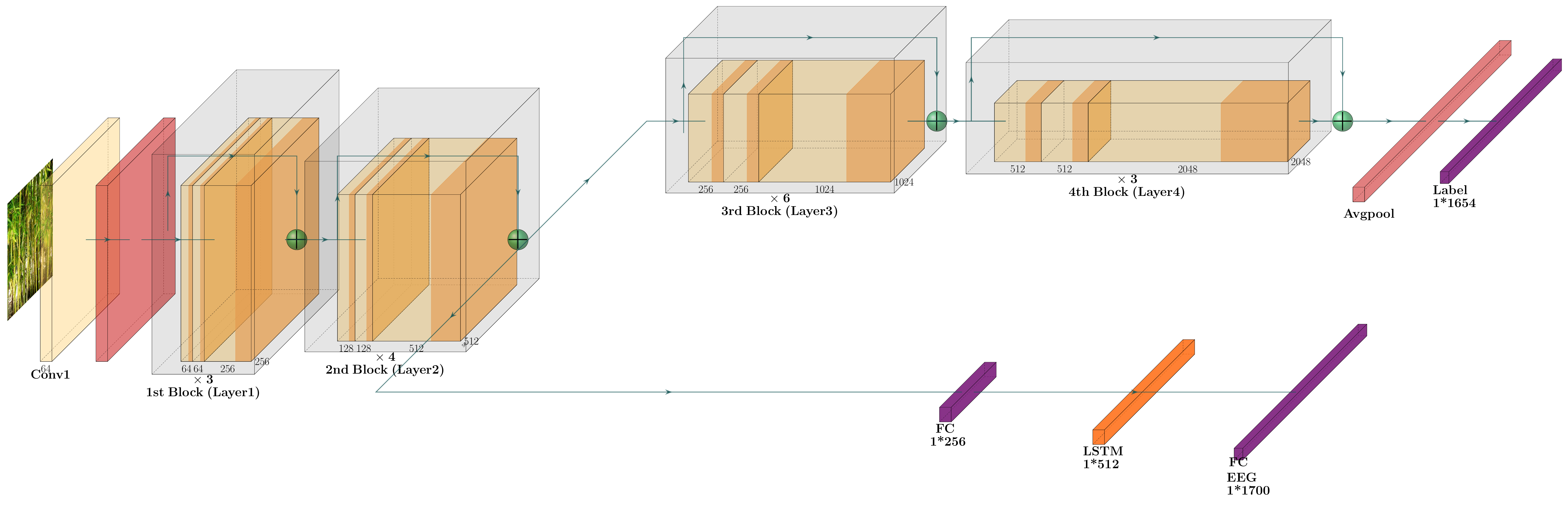} 
        \caption{RNN cluster with long-short-term-memory (LSTM) units. We labeled it with $\textbf{RNN (LSTM)\_Bk2}$. A fully-connected layer was used to transform the \textbf{2rd} ResNet50 \textbf{B}loc\textbf{k} output into a vector of $1 \times 256$, after which a \textbf{LSTM} unit was utilized to map from image features to EEG.}
        \label{fig:bk4_lstm}
       \end{subfigure}

      \begin{subfigure}[t]{0.9\textwidth}
   	\includegraphics[width=10.5cm,height=4.5cm]{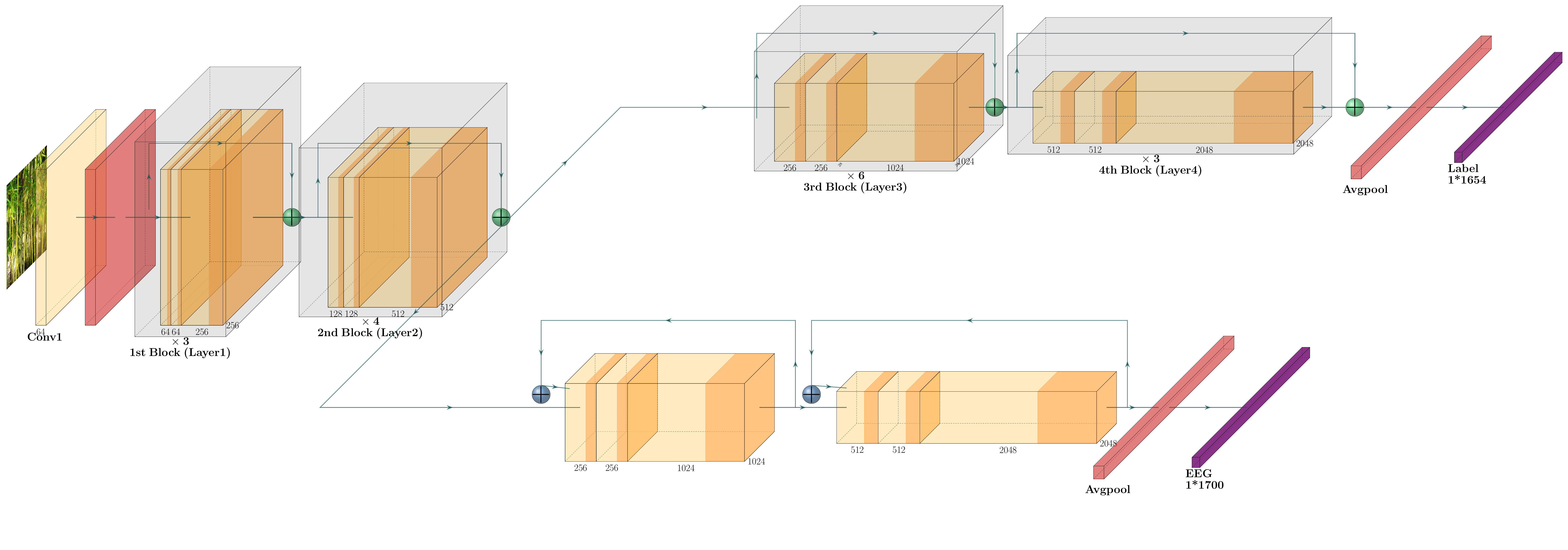} 
        \caption{RNN cluster with ResNet-like feedback connection. We labeled it with $\textbf{RNN\_Bk2}$. The input to the RNN was the \textbf{2rd} ResNet50 \textbf{B}loc\textbf{k} output.}
        \label{fig:bk2_rnn}
      \end{subfigure}

       \begin{subfigure}[t]{0.9\textwidth}
   	 \includegraphics[width=10.5cm,height=1.0cm]{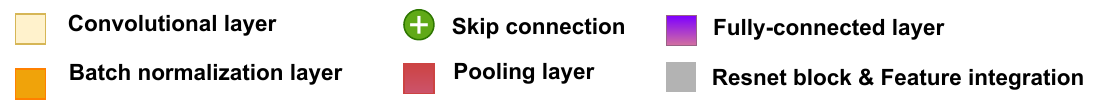} 
          \caption{Legend.}
        \end{subfigure}
     
        \caption{2 architectures in RNN cluster}
        \label{fig:Re_arch}
\end{figure}

\begin{figure}[!htb]
    \centering
    \begin{subfigure}[t]{0.9\textwidth}
       \includegraphics[width=10.5cm,height=4.5cm]{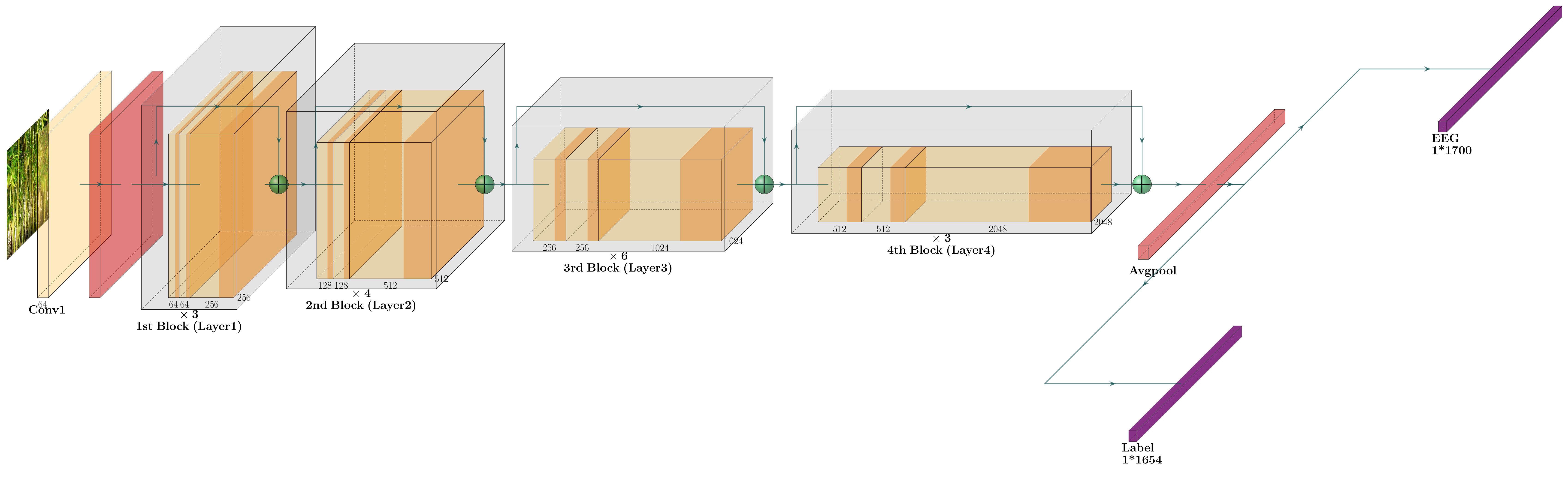} 
       \label{fig:fc_only}
        \caption{Architecture of \textbf{CNN\_Bk4} in CNN cluster. The input to the fully-connected layer was the \textbf{4th} ResNet50 \textbf{B}loc\textbf{k} output.}
     \end{subfigure}	
       
     \begin{subfigure}[t]{0.9\textwidth}
   	\includegraphics[width=10.5cm,height=4.5cm]{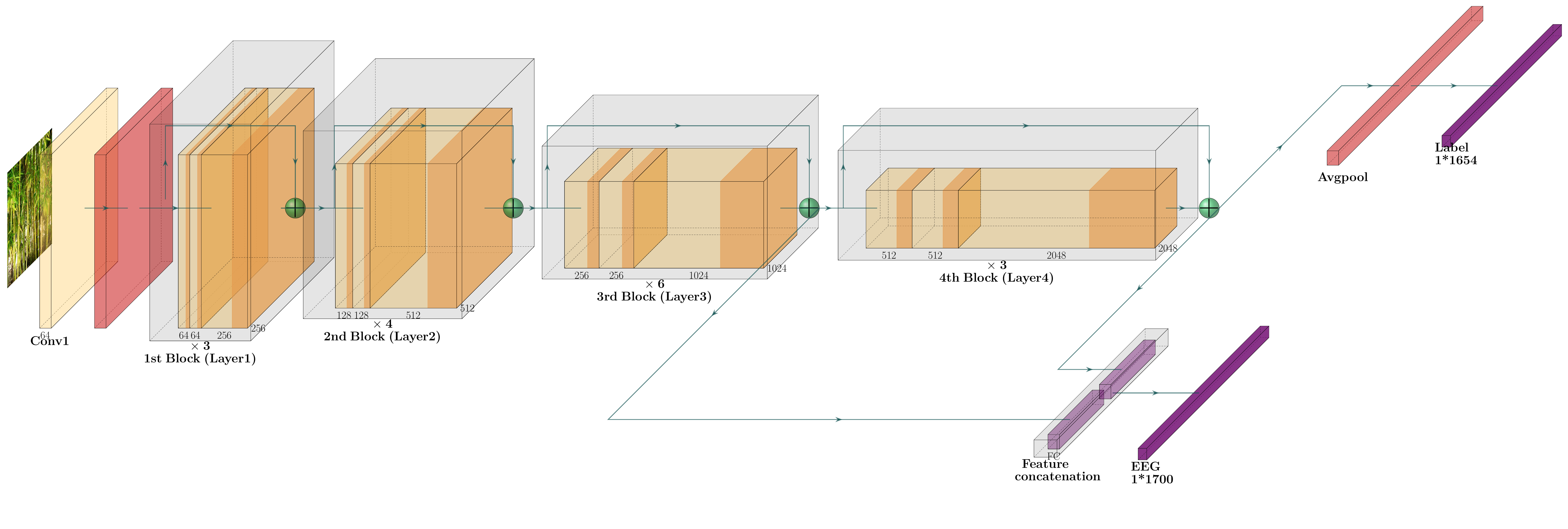} 
        \caption{Architecture of \textbf{CNN\_Bk3,4} in CNN cluster. It first extracted the \textbf{3rd} and \textbf{4th} \textbf{B}loc\textbf{k} features and used 2 fully-connected layers to transform these features dimension to be $1 \times 256$, respectively. The input to the last fully-connected layer for EEG prediction was obtained by \textbf{concat}enating the two transformed features (of dimension: $1 \times 512$) and the output dimension of the last fully-connected layer is $1 \times 1700$ (EEG dimension).}
        \label{fig:conc_bk3_bk4}
        \end{subfigure}

      \begin{subfigure}[t]{0.9\textwidth}
   	\includegraphics[width=10.5cm,height=4.5cm]{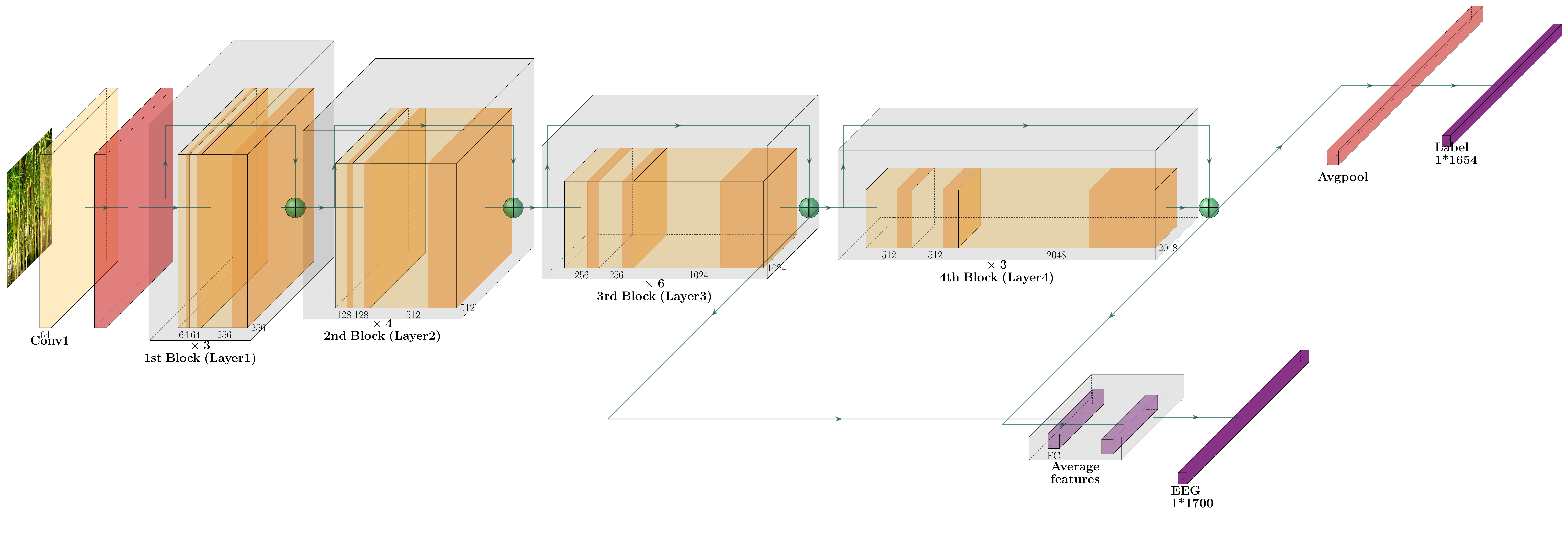} 
        \caption{One architecture similar to \textbf{CNN\_Bk3,4} but averaging the 3rd and 4th block features for EEG prediction. We labeled it as $\textbf{CNN(avg)\_Bk3,4}$. It first extracted the \textbf{3rd} and \textbf{4th} \textbf{B}loc\textbf{k} features and used 2 fully-connected layers to transform these features dimension to be $1 \times 256$, respectively. The input to the last fully-connected layer was obtained by \textbf{av}era\textbf{g}ing the two transformed features (of dimension: $1 \times 256$) and output dimension of the last fully-connected layer is $1 \times 1700$ (EEG dimension).}
        \label{fig:pool_bk3_bk4}
      \end{subfigure}
		
      \caption{3 architectures in CNN cluster.}
      \label{fig:De_arch}
\end{figure}

\begin{figure}[!htb]
   \centering         
    \begin{subfigure}[t]{0.9\textwidth}
      \includegraphics[width=11.5cm,height=5cm]{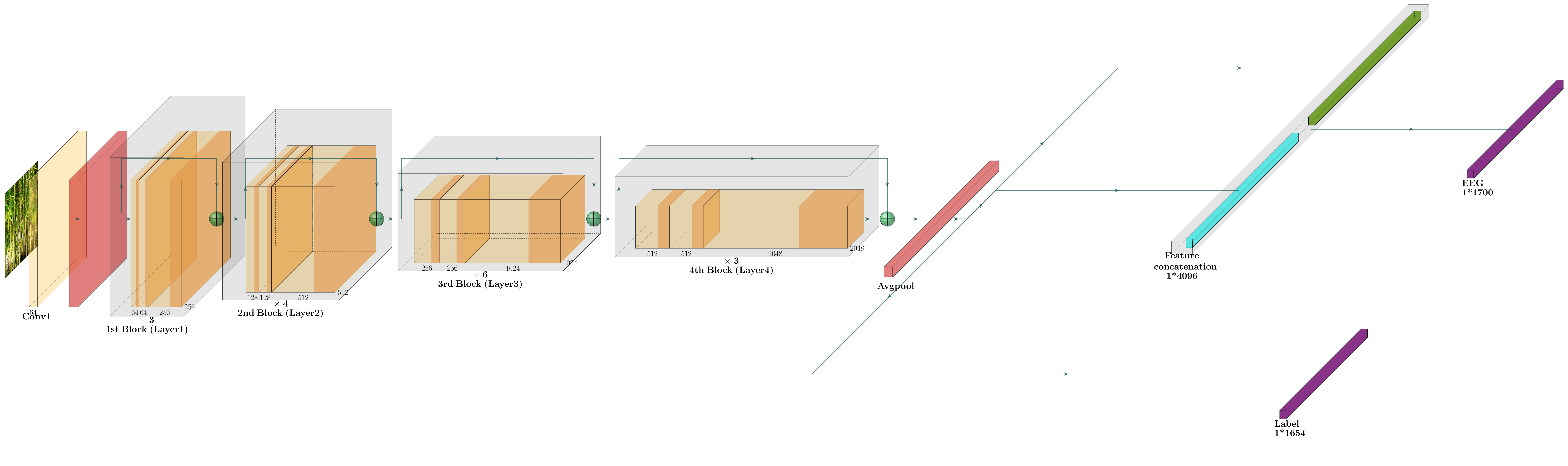} 
      \label{fig:bk4_att}
      \caption{One architecture in Attention layer cluster using position attention module (PAM) and channel attention module (CAM) in the dual attention network. Features extracted from the shared net (last pooling layer in ResNet50) were fed into PAM and CAM modules respectively. The input dimension was $1 \times 2048$ and the output from CAM or PAM was $1 \times 2048$. These features were then concatenated (of dimension: $1 \times 4096$) and fed into a dense layer for EEG prediction (of dimension: $1 \times 1700$). We labeled it as $\textbf{Att\_Bk4}$. The input to independent EEG prediction net was the \textbf{4th} ResNet50 \textbf{B}loc\textbf{k} output.}
    \end{subfigure}

    \begin{subfigure}[t]{0.9\textwidth}
       \includegraphics[width=10.5cm,height=4.5cm]{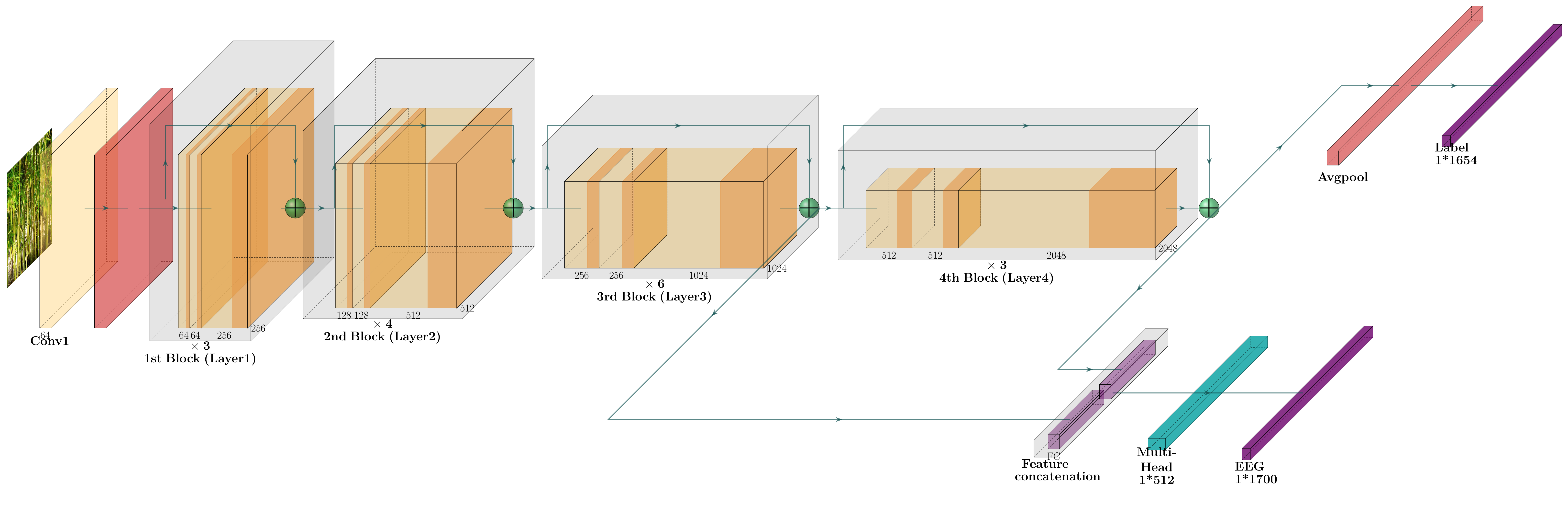} 
       \label{fig:multihead}
       \caption{Another architecture in Attention layer cluster using a multi-head layer to predict EEG. We labeled it with $\textbf{Att(concat)\_Bk3,4}$. The input to the multi-head layer was obtained by \textbf{concat}enating the \textbf{3rd} and \textbf{4th} \textbf{B}loc\textbf{k} (of dimension: $1 \times 512$).}
    \end{subfigure}

    \begin{subfigure}[t]{0.9\textwidth}
       \includegraphics[width=10.5cm,height=4.5cm]{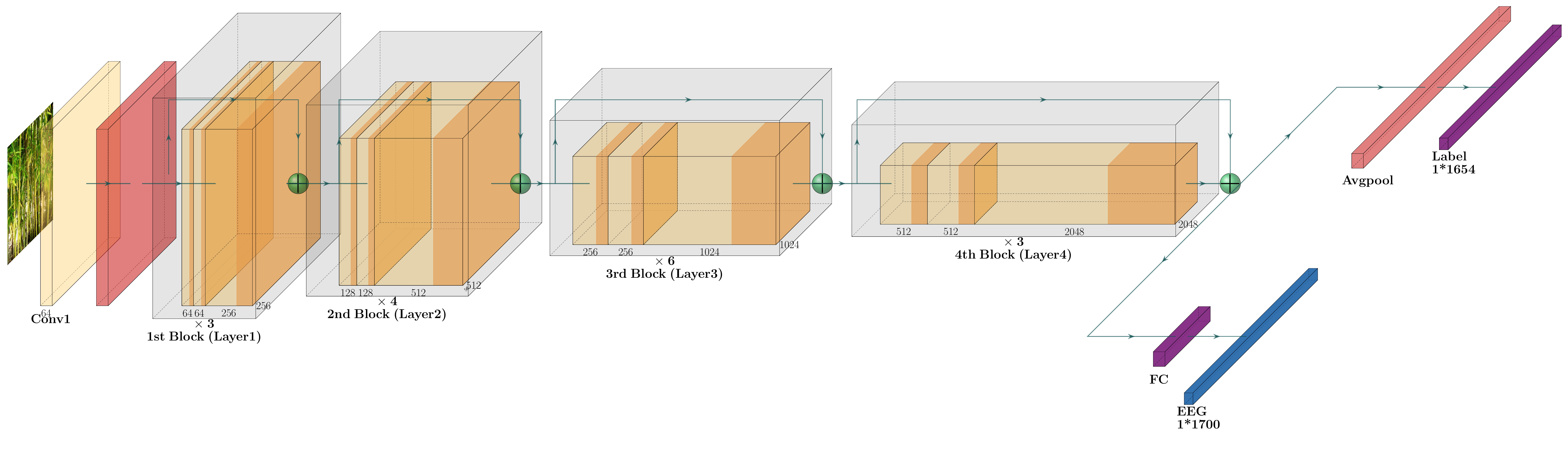} 
       \caption{One architecture in Transformer cluster using Transformer encoders with 6 layers, 32 multi-heads, embedding dimension of 256 and output dimension of 1700. We labeled it as $\textbf{Trans\_Bk4}$. The input to the Transformer encoder was \textbf{4th} \textbf{B}loc\textbf{k} output.}
       \label{fig:trans}
    \end{subfigure}

    \caption{2 architectures in Attention layer cluster and 1 architecture in Transformer.}
    \label{fig:Te_Att_arch}
\end{figure}

\paragraph{Architecture}
In CNN cluster, we concatenated or averaged output from different ResNet50-blocks in the shared net and then used fully connected layers to map from image features to EEG signals directly. In RNN cluster, shallow RNNs where feedback were realized with Resnet-like skip connections, or Long short-term memory (LSTM) units were integrated with the shared net, as recurrence is capable of capturing temporal dynamics and patterns in time-series data. In Transformer cluster, 6 layers of transformer encoders with 32 multi-heads, dropout value of 0.5 and embedding dimension of 256 were combined with the shared net, as transformer architectures have been shown to be beneficial for fMRI and MEG prediction \cite{schwartz2019inducing}. In Attention layer cluster, 2 self-attention layers were considered. One is a self-attention layer in \cite{vaswani2017attention}. We used 4 multi-heads and an embedding dimension of 256 after optimization experiments. Another utilized the position attention modules (PAM) and channel attention modules (CAM) used in \cite{fu2020scene}. We followed the default settings. All the 24 architectures from the 4 clusters are included in \Cref{tab:tab1}. Some architectures from the 4 clusters are depicted in \Cref{fig:De_arch}, \Cref{fig:Re_arch} and \Cref{fig:Te_Att_arch}.

\FloatBarrier

\FloatBarrier

\FloatBarrier

\FloatBarrier
\paragraph{Model training}
\cite{kendall2018multi} used the total loss function in Equation\Cref{eq:uncertainty} to balance the loss updating during dual task learning. 
\begin{equation}
\label{eq:uncertainty}
L(\textbf{W}, \delta_1, \delta_2) = \frac{1}{2 \delta_1^2} L_1(\textbf{W}) + \frac{1}{2 \delta_2^2} L_2(\textbf{W}) + log \delta_1 +log \delta_2
\end{equation}
Where $L_1$ and $L_2$ were EEG prediction and image classification loss, respectively. $\frac{1}{2 \delta_1^2}$ and $\frac{1}{2 \delta_1^2}$ were loss coefficients. $\frac{1}{2 \delta_1^2}$, $\frac{1}{2 \delta_1^2}$, $L_1$ and $L_2$ were updated using Adam optimizer, with a learning rate of 5e-6 and a weight decay of 0.0. The learning rate and weight decay values were optimized with cross validation. \textbf{W} denoted weights of model. The image classification branch was pre-trained on Imagenet and the independent net of EEG prediction branch were initialized with 3 different pre-chosen training seeds (0, 17 and 337).

\subsection{Evaluation of EEG prediction and adversarial robustness}
\label{sec_app:pgd&eeg}

\paragraph{EEG prediction evaluation}

Given the hierarchical visual processing in the brain, EEG signals at different time points may capture activities in different regions. We used Pearson correlation analysis to evaluate EEG predictions at various time points using 1654 images from the validation set. The Pearson correlation coefficient (PCC) at each time point was computed as follows: (1) By looping through the channel and temporal dimension, we measured the PCC between predicted EEG and biological EEG at different channel index (ci) and time point\textbf{s}. Consequently, we obtained a PCC array of size $17 \times 100 $, denoted as $PCC(ci,tps)$. (2) We averaged the $PCC(ci,tps)$ across the channel dimension and obtained $PCC(pts)$, which denoted linear relationship between predicted and biological EEG at different time points. Here, $pts \in [-0.02s,-0.01s,...,0.07s,0.08s]$. The formula for PCC was given as follows: $PCC = \frac{cov(signal\_1,signal\_2)}{\sigma_1,\sigma_2}$ , where $cov$ denoted the covariance while $\sigma_1$ and $\sigma_2$ denoted the standard deviation of $signal\_1$ and $signal\_2$.

\paragraph{Adversarial examples generated with PGD}

Projected Gradient Descent, or PGD, iteratively constructs adversarial examples as follows : 

\begin{equation}
x^{t+1} = Proj_{x+S}(x^t + \alpha sgn(\nabla_x L(\theta,x^t,y)))
\end{equation}

where $\theta$ is the parameters of models, $x^t$ is the input and y is the associated label. $L(\theta,x^t,y)$ is the loss of model training and $\nabla_x L(\theta,x^t,y)$ is the gradient of loss $L$ with respect to input $x$. $\alpha$ is the gradient step size. t is the number of steps for iteration. $x^{t}$ and $x^{t+1}$ are adversarial examples before and after next iteration, respectively. The $\emph{Proj}$ is an operator to construct $x^{t+1}$ within space $x+S$, such as $l_\infty$ ball or $l_2$ norm ball around x. In our setting, t was 50 and 40 for $l_2$-bounded PGD and $l_\infty$-bounded PGD, respectively. In $l_\infty$-constrained PGD, the attack strength $\epsilon\in [1e^{-5},2e^{-5},3e^{-5},4e^{-5},5e^{-5},6e^{-5},7e^{-5},8e^{-5},1e^{-4},3e^{-4},5e^{-4},7e^{-4},8e^{-4}\\,1e^{-3},8e^{-3},1e^{-2}]$ and the step size relative to $\epsilon$ was $0.01/0.3$. In the $l_2$-constrained PGD, the attack strength $\epsilon\in [1e^{-3},5e^{-3}, 7e^{-3}, 1e^{-2}, 2e^{-2}, 3e^{-2}, 5e^{-2}, 7e^{-2}, 1e^{-1}\\, 2e^{-1}, 3e^{-1}, 5e^{-1}, 7e^{-1},1.0]$ and the step size relative to $\epsilon$ was set to 0.025.

\paragraph{Adversarial examples generated with  Carlini \& Wagner ($C\&W $) attack}
$C\&W $ attack formulates the generation of adversarial examples as an optimization problem, finding the smallest perturbations to the input data that causes misclassification of the target model. $C\&W $ attack defines the objective function $J(x^{'})$ as follows:
\begin{equation}
J(x^{'}) = \alpha \cdot dist(x,x^{'}) + \beta \cdot loss(f(x^{'}),y)
\end{equation}
Where $x$ is the original image; $x^{'}$ is the perturbed image; In the case of L2 $C\&W $ attack, $dist(x,x^{'})$ measures the perturbation using the $l_2$ norm. $loss(f(x^{'}),y)$ represents the misclassification loss of target model $f$ on the perturbed input with respect to the target class y. $\alpha$ and $\beta$ are weights to balance the $dist(x,x^{'})$ and $loss(f(x^{'}),y)$. The $C\&W $ attack iteratively adjusts the perturbations to improve the chances of misclassification while keeping the perturbations imperceptible. Thus the term $dist(x,x^{'})$ is minimized and $loss(f(x^{'}),y)$ is maximized. Gradient descent is used for optimization. The perturbation is updated until it converged towards an examples $x^{'} = x^{'} - \eta \cdot \nabla_x^{'}J(x^{'})$, where $\eta$ is the step size. The attack strength $\epsilon\in [1e^{-5},7e^{-5},1e^{-4},7e^{-4},1e^{-3},1e^{-2},1e^{-1},3e^{-1},5e^{-1},7e^{-1},\\9e^{-1},1.0,1.2,1.4,1.6,1.8,2.0,2.2,2.4,2.6,2.8,3.0]$ and the step size relative to $\epsilon$ was $0.01$. 

\subsection{Mean adversarial robustness gain and mean EEG prediction accuracy}
\label{sec_app:mean_gain_pcc}
Mean adversarial robustness gain and mean EEG prediction accuracy were used to measure the relationship between adversarial robustness gain and EEG prediction accuracy. We trained 24 architectures from all 4 clusters on EEG of 10 subjects with 3 training seeds. As a result, we had totally $24 \times 10 \times 3 = 720$ sample models for analysis. 

\paragraph{Mean adversarial robustness gain ${Avg\_Gain_{DTL}}$}
For each sample model n ($0 \leq n \leq 720$), we first computed $Avg\_Gain_{DTL}^n$ which was the averaged value of $Gain_{DTL}^n(\epsilon)$ across selected attack strength. In C\&W, $Gain_{DTL}(\epsilon)$ used for $Avg\_Gain_{DTL}^n$ calculation included those obtained under attack strength $\epsilon\in [5e^{-1},7e^{-1},9e^{-1},1.0,1.2,1.4,1.6,1.8,2.0,2.2,2.4,2.6,2.8,3.0]$. In $l_\infty$-constrained PGD, $Gain_{DTL}(\epsilon)$ used for $Avg\_Gain_{DTL}^n$ calculation included those obtained under attack strength $\epsilon\in [8e^{-5},1e^{-4},3e^{-4},5e^{-4}, 7e^{-4},8e^{-4},1e^{-3},8e^{-3},1e^{-2}\\,1e^{-1}]$. In $l_2$-constrained PGD, we considered the attack strength $\epsilon\in [7^{e-2},1e^{-1}\\,2e^{-1},3e^{-1},5e^{-1},7e^{-1},1.0]$. We selected high $\epsilon$ values because under high attack strength, robust models would perform much better than fragile ones in classifying adversarial examples, which enabled us to better quantify relationship between EEG prediction accuracy and robustness gain. We finally had 720 $Avg\_Gain_{DTL}^n$ values and the ${Avg\_Gain_{DTL}}$ was computed by averaged $Avg\_Gain_{DTL}^n$ across 3 training seeds and 10 subjects.


\paragraph{Mean EEG prediction accuracy $Avg\_PCC\_tps$ and optimal sliding window selection}
Based on our initial experimentation, we posited that higher prediction accuracy of EEG signals around 100ms might result in increased gains in adversarial robustness. To rigorously investigate this association, we initially deployed three sliding windows covering the vicinity of 100ms: 0.10s to 0.12s, 0.09s to 0.14s, and 0.05s to 0.3s, respectively. Within these windows, for each model iteration (n), we computed Pearson correlation coefficient (PCC) values ($PCC^n(pts)$) as described in \Cref{sec_app:pgd&eeg}, subsequently averaging these values ($Avg\_PCC^n$) across all time points within the sliding window. By averaging $Avg\_PCC^n$ values across 10 subjects and 3 training seeds, we derived the ${Avg\_PCC\_tps}$. Assessing the correlation between ${Avg\_Gain_{DTL}}$ and different ${Avg\_PCC\_tps}$ obtained with the three window sizes, we identified an optimal window size of 0.06s, resulting in the highest correlation value between ${Avg\_Gain_{DTL}}$ and ${Avg\_PCC\_tps}$ around 100ms. Subsequently, we proceeded to identify the most informative EEG signals crucial for robustness gain by moving this optimal sliding window across all time points, with a step size of 0.01s, and measuring the correlation values between ${Avg\_Gain_{DTL}}$ and $Avg\_PCC\_tps$ across all time points. In total, we obtained 100 correlation values, which represent the correlation between ${Avg\_Gain_{DTL}}$ and ${Avg\_PCC\_tps}$ when the optimal sliding window arrives at different time points. Similarly, to discern the most significant channels for enhancing the robustness gain, after identifying these significant time points, we averaged ${PCC(ci,tps)}$ across critical time points, yielding the averaged prediction accuracy across critical time points for each channel. 

\clearpage
\section{Additional results}
\label{sec_app:add_re}

\subsection{Relationship between $Avg\_Gain_{DTL}$ and $Avg\_PCC\_tps$ within different sliding windows}
\label{sec_app:relation_2_slide}

\begin{figure}[!htb]
  \centering
  \includegraphics[height=5.2cm]{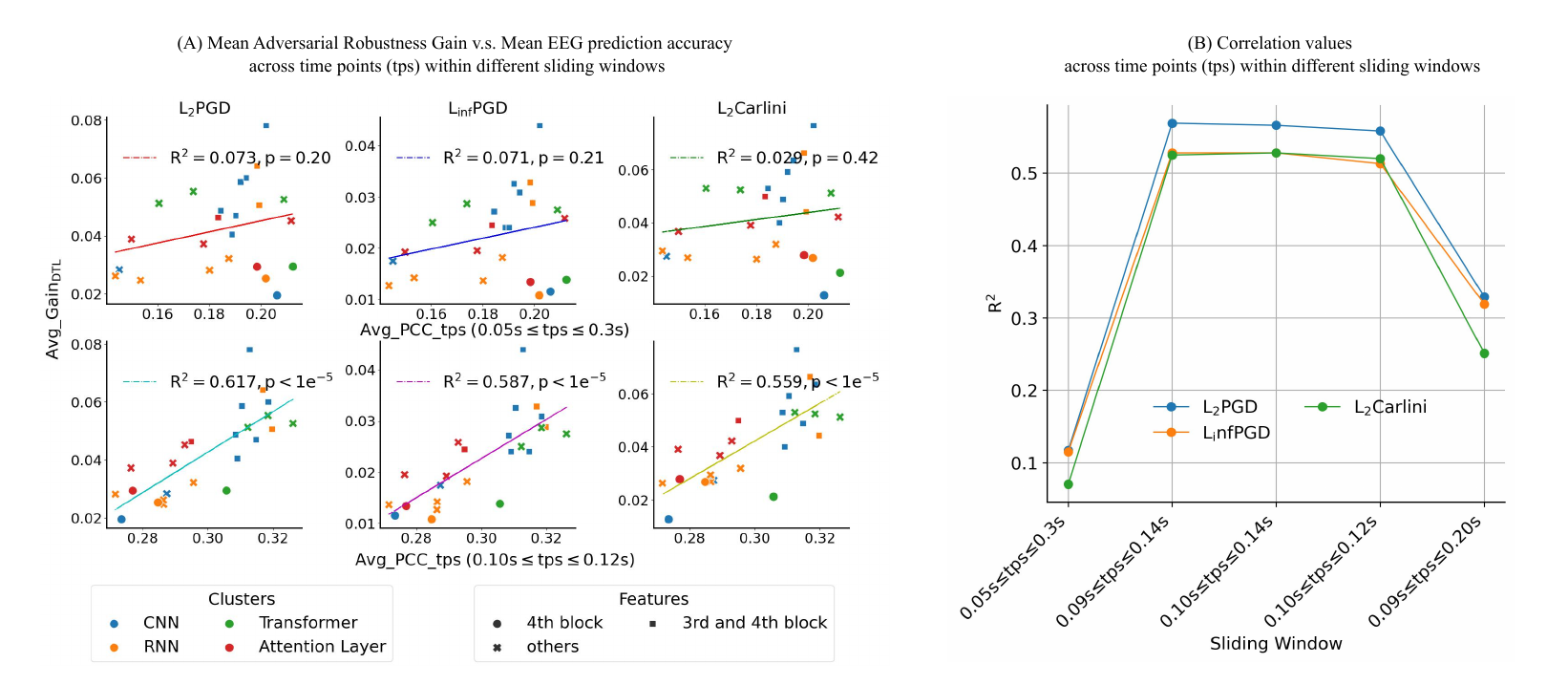}
  \caption{Correlation between $Avg\_Gain_{DTL}$ and $Avg\_PCC\_tps$ across various sliding windows. (A) shows how $Avg\_Gain_{DTL}$ varies with $Avg\_PCC\_tps$ within two specific sliding windows: one ranging from 0.05s to 0.3s, and the other from 0.10s to 0.12s. The correlation values observed within these windows are lower compared to those depicted in \Cref{fig:corr_slide} (B). Here, (B) presents the $R^2$ values for several sliding windows, indicating that the window ranging from 0.09s to 0.14s achieves the highest $R^2$ value. This sliding window is also identified as optimal in \Cref{fig:corr_slide} (B). 
  }
  \label{fig:opt_slid}
\end{figure}

In \Cref{fig:opt_slid} (A), the correlation values between $Avg\_Gain_{DTL}$ and $Avg\_PCC\_tps$ within two specific sliding windows, one ranging from 0.05s to 0.3s and the other from 0.10s to 0.12s, were inferior to those in \Cref{fig:corr_slide} (B), which suggested that using a sliding window size of 0.06s allowed us to capture more significant EEG signals around 100ms for adversarial robustness improvements. Using this optimal window size, \Cref{fig:corr_slide} rigorously examines the correlation between EEG prediction accuracy and adversarial robustness gain. 

\begin{table}[tb]
\centering
\caption{All architectures.}
\label{tab:tab1}
\resizebox{\textwidth}{!}{
\begin{threeparttable}
\begin{tabularx}{\textwidth}{|l|l|X|l|l|l|}
\hline
Model & Cluster & Name & \multicolumn{3}{c|}{(Avg\_PCC\_tps, Avg\_Gain)} \\
\hline
 & & & $L_2$ PGD & $L_\infty$ PGD & $L_2$ Carlini \\
\hline
0  & CNN  & $\texttt{CNN\_Bk4}$ & (0.264,0.019)& (0.264,0.012) & (0.264,0.013)\\
\hline
1  & CNN  & $\texttt{CNN(concat)\_Bk12}$ & (0.267,0.028)& (0.267,0.017) & (0.267,0.027)\\
\hline
2  & CNN  & $\texttt{CNN(concat)\_Bk1234}$ \tnote{1} & (0.290,0.05) & (0.290,0.027) & (0.290,0.05) \\
\hline
3  & CNN  & $\texttt{CNN(avg)\_Bk1234}$ \tnote{2} & (0.292,0.04) & (0.292,0.024)& (0.292,0.04) \\
\hline
4  & CNN  & $\texttt{CNN(concat)\_Bk34}$ & (0.296,0.08)& (0.296,0.044) & (0.296,0.076)\\
\hline
5  & CNN  & $\texttt{CNN(avg)\_Bk34}$ & (0.293,0.06) & (0.293,0.033) & (0.293,0.059) \\
\hline
6  & CNN  & $\texttt{CNN(concat)\_Bk234}$ & (0.299,0.060) & (0.299,0.031) & (0.299,0.063)\\
\hline
7  & CNN  & $\texttt{CNN(avg)\_Bk234}$ & (0.296,0.047) & (0.296,0.02) & (0.296,0.049) \\
\hline
8  & RNN  & $\texttt{RNN(concat)\_Bk34}$ & (0.30,0.06) & (0.30,0.033) & (0.30,0.066)\\
\hline
9  & RNN  & $\texttt{RNN(avg)\_Bk34}$ & (0.302,0.051) & (0.302,0.029) & (0.302,0.04)\\
\hline
10 & RNN  & $\texttt{RNN(LSTM)\_Bk2}$ \tnote{3} & (0.27,0.025) & (0.27,0.014) & (0.27,0.027)\\
\hline
11 & RNN  & $\texttt{RNN(LSTM)\_Bk1}$ & (0.267,0.026) & (0.267,0.013) & (0.267,0.029)\\
\hline
12 & RNN  & $\texttt{RNN\_Bk4}$ & (0.271,0.025) & (0.271,0.011) & (0.271,0.027) \\
\hline
13 & RNN  & $\texttt{RNN\_Bk2}$ & (0.279,0.032) & (0.279,0.018) & (0.279,0.032) \\
\hline
14 & RNN  & $\texttt{RNN\_Bk3}$ & (0.258,0.028) & (0.258,0.014) & (0.258,0.026) \\
\hline
15 & \texttt{Trans}former & $\texttt{Trans\_Bk4}$ & (0.291,0.029) & (0.291,0.014) & (0.291,0.021)\\
\hline
16 & \texttt{Trans}former & $\texttt{Trans\_Bk1}$ & (0.289,0.051) & (0.289,0.025) & (0.289,0.053)\\
\hline
17 & \texttt{Trans}former & $\texttt{Trans\_Bk2}$ & (0.297,0.055) & (0.297,0.029) & (0.297,0.052)\\
\hline
18 & \texttt{Trans}former & $\texttt{Trans\_Bk3}$ & (0.309,0.053)& (0.309,0.03) & (0.309,0.051)\\
\hline
19 & \texttt{Att}ention layer & $\texttt{Att(concat)\_Bk34}$ & (0.278,0.046) & (0.278,0.024) & (0.278,0.05)\\
\hline
20 & \texttt{Att}ention layer & $\texttt{Att(concat)\_Bk12}$ & (0.271,0.039) & (0.271,0.019) & (0.271,0.037)\\
\hline
21 & \texttt{Att}ention layer & $\texttt{Att\_Bk4}$ & (0.265,0.029) & (0.265,0.013) & (0.265,0.028)\\
\hline
22 & \texttt{Att}ention layer & $\texttt{Att\_Bk3}$ & (0.283,0.045) & (0.283,0.026) & (0.283,0.042)\\
\hline
23 & \texttt{Att}ention layer & $\texttt{Att\_Bk2}$ & (0.264,0.037) & (0.264,0.02) & (0.264,0.039)\\
\hline
\end{tabularx}
\begin{tablenotes}
\item[1] Represents architecture in \textbf{CNN} cluster that \textbf{concat}enates the output from \textbf{B}loc\textbf{k}s 1, 2, 3, and 4 of ResNet50 for EEG prediction. Model 2 and model 5 exhibited similar $Avg\_PCC\_tps$ and $L_2$ Carlini \& Wagner $Avg\_Gain_{DTL}$. In the third subplot of \Cref{fig:corr_slide} B, two blue squares representing these two models overlapped. 
\item[2] Represents architecture that \textbf{av}era\textbf{g}es the output from \textbf{B}loc\textbf{k}s 1, 2, 3, and 4 of ResNet50 for EEG prediction.
\item[3] Represents architecture that predicts EEG from \textbf{B}loc\textbf{k} 2 features using \textbf{LSTM} units.
\end{tablenotes}
\end{threeparttable}}
\end{table}

\end{document}